\documentclass[runningheads]{llncs}

\usepackage[T1]{fontenc}
\usepackage{graphicx,verbatim}
\usepackage{amsmath}
\usepackage{amsfonts}
\usepackage{multirow}
\usepackage{graphicx}
\usepackage[normalem]{ulem}
\usepackage[table,xcdraw]{xcolor}
\usepackage{makecell}
\usepackage{amssymb}% http://ctan.org/pkg/amssymb
\usepackage{pifont}% http://ctan.org/pkg/pifont
\newcommand{\xmark}{\ding{53}}%
\useunder{\uline}{\ul}{}
\usepackage{orcidlink}
\usepackage[misc]{ifsym}
\usepackage{hyperref}
\usepackage{color}

\usepackage{booktabs}
\usepackage{arydshln}

\begin{document}

\title{Interpretable Rheumatoid Arthritis Scoring via Anatomy-aware Multiple Instance Learning}
\titlerunning{Anatomy-aware Deep Learning for Rheumatoid Arthritis Scoring}

\author{Zhiyan Bo\inst{1}\textsuperscript{(\Letter)}\orcidlink{0009-0002-6458-3156} \and
Laura C. Coates\inst{2}\orcidlink{0000-0002-4756-663X} \and
Bart\l omiej W. Papie\.z\inst{1}\orcidlink{0000-0002-8432-2511}}

% 1{Bo, Zhiyan} 
% 2{Coates, Laura C.} 
% 3{Papiez, Bartlomiej W.} 

\authorrunning{Z. Bo et al.}
\institute{Big Data Institute, Nuffield Department of Population Health, University of Oxford, Oxford, UK \\
\email{zhiyan.bo@reuben.ox.ac.uk} \&
\email{bartlomiej.papiez@bdi.ox.ac.uk}\\
\and
Nuffield Department of Orthopaedics, Rheumatology and Musculoskeletal Sciences, University of Oxford, Oxford, UK}

\maketitle              
\begin{abstract}

The Sharp/van der Heijde (SvdH) score has been widely used in clinical trials to quantify radiographic damage in Rheumatoid Arthritis (RA), but its complexity has limited its adoption in routine clinical practice. To address the inefficiency of manual scoring, this work proposes a two-stage pipeline for interpretable image-level SvdH score prediction using dual-hand radiographs. Our approach extracts disease-relevant image regions and integrates them using attention-based multiple instance learning to generate image-level features for prediction. We propose two region extraction schemes: 1) sampling image tiles most likely to contain abnormalities, and 2) cropping patches containing disease-relevant joints. With Scheme 2, our best individual score prediction model achieved a Pearson's correlation coefficient (PCC) of 0.943 and a root mean squared error (RMSE) of 15.73. Ensemble learning further boosted prediction accuracy, yielding a PCC of 0.945 and RMSE of 15.57, achieving state-of-the-art performance that is comparable to that of experienced radiologists (PCC = 0.97, RMSE = 18.75). Finally, our pipeline effectively identified and made decisions based on anatomical structures which clinicians consider relevant to RA progression. 

\keywords{Rheumatoid Arthritis  \and Hand X-ray scoring }
\end{abstract}

\section{Introduction}
Rheumatoid Arthritis (RA) affects around 18 million people worldwide~\cite{WHO}. This autoimmune disease causes joint inflammation, ultimately leading to structural damage and disability. Commonly manifesting joint erosion and joint space narrowing (JSN) in hands, wrists, and feet, RA is evaluated by plain radiography for diagnosis and monitoring~\cite{BMJ}.
Several RA quantification schemes exist across imaging modalities, with the van der Heijde modification of the Sharp (SvdH) score being the standard for radiography in clinical trials due to its strong intra- and inter-observer reliability and sensitivity to changes~\cite{Landewe2016,Salaffi2019}. SvdH examines 16 areas and 15 joints in each hand and wrist (see Fig.~\ref{fig_pipelines_overview}(A)) and 6 areas and 6 joints in each foot~\cite{VanDerHeijde1999}. However, its detailed evaluation requires approximately 25 minutes per patient by an experienced radiologist, limiting its practicality in clinical settings~\cite{Boini2001}. In addition, suboptimal imaging conditions and superimposition contribute to inter-observer variability, particularly in detecting small changes, posing challenges on clinical trials in early-stage patients~\cite{Landewe2016}. 

Deep learning (DL) provides a possibility to reduce reliance on human scorers for RA quantification while improving sensitivity, consistency, and efficiency~\cite{Momtazmanesh2022}. Several automated RA scoring methods have been proposed for ultrasound and X-ray imaging~\cite{Bird2022,Gilvaz2023}, typically consisting of a joint localisation stage (using U-Net-based heatmap regression, YOLO model or Mask-RCNN) followed by convolutional neural network (CNN)-based joint-level damage quantification~\cite{Bird2025,Hemalatha2019,Ho2018}. 
Some methods focused on specific anatomical structures, such as fingers \cite{Hirano2019}, making them insufficient for holistic SvdH scoring. The shortage of datasets with joint-level annotations additionally increases the difficulty of method validation and comparison. A few methods developed using in-house datasets~\cite{Honda2023} lacked image-level or patient-level performance evaluation. Most methods covering both hand and foot joints relied on 674 sets of images from the RA2-DREAM challenge~\cite{Sun2022}, but since participants were not granted access to the final evaluation images, detailed validation was limited~\cite{Maziarz2022,Tan2021}. External validation was performed only in \cite{Venäläinen2024}, where the proposed pipeline and the top two solutions of the challenge were tested on a private dataset of 205 patients. 

While joint-level scoring provides detailed assessments, accurately localising individual joints can be challenging in late-stage RA patients, who may struggle to straighten deformed fingers during scans. Also, the intrinsic ambiguity between adjacent scores increases the task difficulty, potentially leading to large radiograph-level error~\cite{Bird2025}.
To address this, image-level scoring methods have been explored such as an overall SvdH score prediction for dual-hand radiographs~\cite{Bo2024,Wang2022}. In~\cite{Moradmand2025}, a four-stage method with image reorientation, hand segmentation, joint identification, and image-level score prediction was developed. Despite promising performance, these methods lack interpretability. While joint patches were used as inputs \cite{Moradmand2025}, the pipeline could not pinpoint damage sites. In our previous work~\cite{Bo2024}, Gradient-weighted Class Activation Mapping (Grad-CAM) was performed to provide explanations for the model's predictions, but our results revealed model confusion by irrelevant regions, such as finger bone diaphyses and arms.

To address these challenges, we propose an anatomy-aware multiple instance learning (MIL) neural network that intelligently extracts RA-relevant patches from dual-hand radiographs and effectively integrates them for total SvdH score prediction. To enable the extraction of regions assessed by SvdH, we also introduce a novel dual-hand joint localisation model. Finally, our MIL framework incorporates an attention mechanism, enhancing both interpretability and predictive performance. Our model achieved state-of-the-art (SOTA) results on a publicly available dataset and performed comparably to experienced radiologists. Additionally, clinician evaluation highlighted the improved clinical relevance of our method in identifying RA-damaged regions.

\section{Methods}
\begin{figure}[!t]
\centering
\includegraphics[width=1\textwidth]{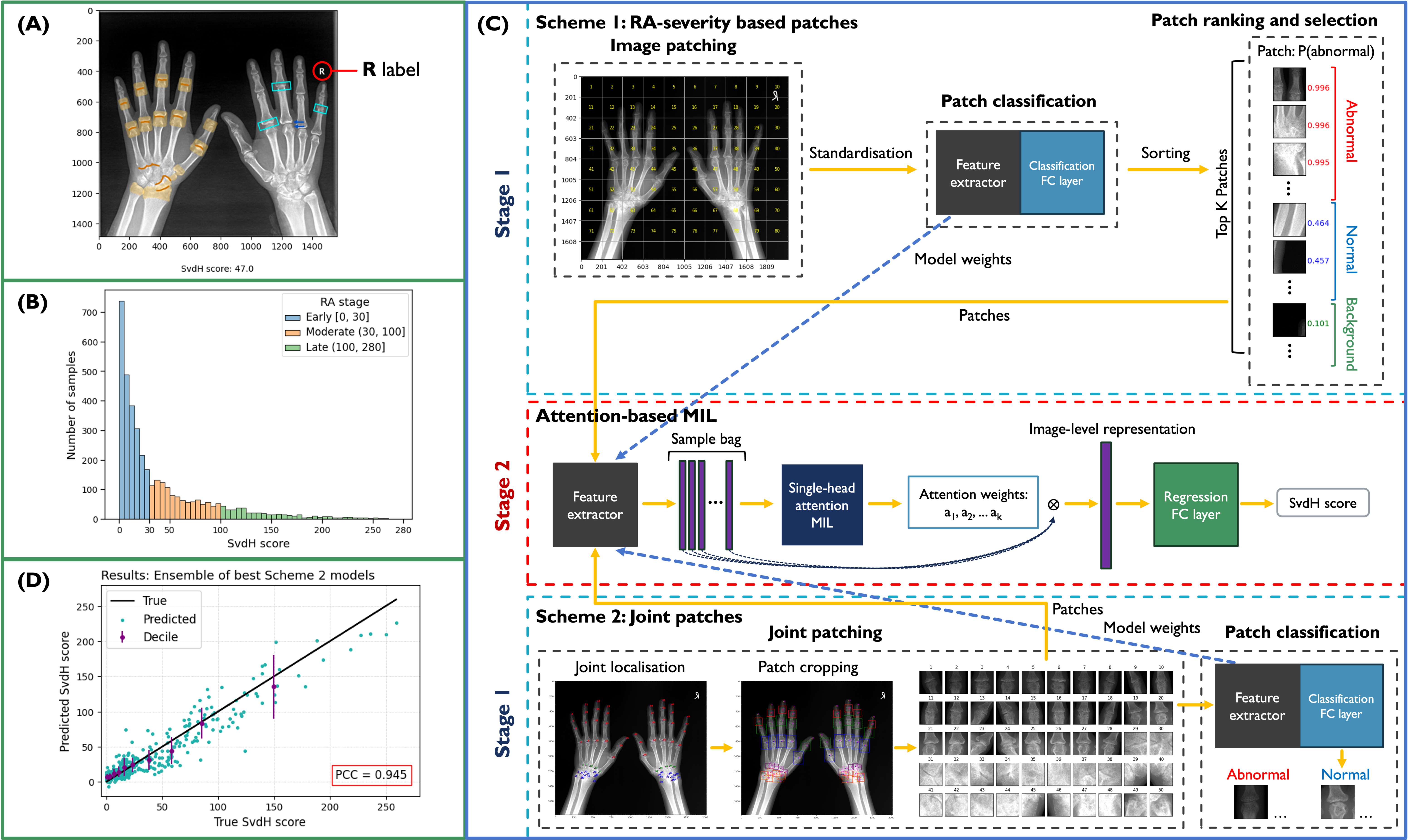}
\caption{(A) Example radiograph from \cite{Wang2022} highlighting bones (yellow) and joints (orange) assessed by SvdH in the left hand and wrist with examples of bone erosion (blue arrows) and JSN (cyan boxes) in the right hand. (B) SvdH score distribution. (C) Overview of the proposed pipelines. In Scheme 2, 37 landmarks are labelled in each hand, which are then used to crop 25 patches. (D) Predicted vs. true SvdH scores for our best score prediction model.} \label{fig_pipelines_overview}
\end{figure}

The proposed framework has two main stages: 1) RA-relevant area sampling and 2) score prediction (see Fig.~\ref{fig_pipelines_overview}(C)). For Stage 1, we propose two strategies: Scheme 1: in which the most "abnormal" tiles in an image are sampled based on RA relevance, and Scheme 2: in which joint patches are cropped to exclude non-relevant anatomical information. In Stage 2, features extracted from the sampled patches are fed into a single-head Attention-Based MIL (ABMIL) model for total SvdH score prediction.

\subsection{Scheme 1: RA-severity based patches}
Scheme 1 was inspired by the intelligent sampling method proposed by \cite{Su2022}. Here, images are divided into non-overlapping square patches, each with a side length of $\frac{1}{10}$ of the image's longer dimension. Then, a weakly supervised classifier categorises patches as normal, abnormal, or background, sorts them by abnormality probability, and samples the top $K$ patches in the order (abnormal, normal, background) prioritising RA-relevant regions. Since patch-level labels were unavailable, we used image-level labels as proxies with images having a total SvdH score $<$ 5 considered normal, and those with a score $\geq$ 70 considered abnormal. To identify background patches, a nnU-Net \cite{Isensee2020} was trained using high-quality segmentation masks created by morphological operations (blurring, erosion, dilation, thresholding, and removal of small bright/dark areas). Then segmentation masks were generated for all images involved in patch classifier (PC) development, with small bright areas removed during post-processing. Patches with $\leq$ 2\% of foreground were labelled as background. Following~\cite{Bo2024,Wang2022}, MobileNetV2~\cite{Sandler2018}, ResNet-34, and ResNet-50~\cite{He2016} were selected as the backbone architectures for PC, with the final fully connected (FC) layer modified to output 3 classes. The PCs were later truncated before the FC layer forming feature extractors (FEs) which output 1280, 512, and 2048 features respectively.

\subsection{Scheme 2: Joint patches}
Joint localisation is a key step of hand radiographic analysis, typically involving identifying joints as landmarks or drawing bounding boxes. Here we defined 37 joint landmarks per hand and wrist (see Fig.~\ref{fig_pipelines_overview}(C)) and cropped patches accordingly to accommodate variations in imaging protocols. To the best of our knowledge, no existing joint detector is designed for dual-hand radiographs, and severe hand deformities in RA patients prevent straightforward separation of images into left and right hands. To address this, we labelled a subset of the dataset~\cite{Wang2022} for all landmarks and adapted the multiresolution Hybrid Transformer-CNN (HTC), a top-performing landmark localisation model originally designed for single-hand radiographs of children~\cite{Viriyasaranon2023}. The number of output channels was modified to 74 to reflect dual-hand radiographs. Using the predicted landmarks, images were aligned to a standard position (fingers pointing upwards) and 25 patches covering most hand joints (see Fig.~\ref{fig_pipelines_overview}(C)) were extracted per side. MobileNetV2, ResNet-34, and ResNet-50 were trained to classify joint patches as normal or abnormal using the same PC-development subset as Scheme~1 and then truncated before the FC layer to form Scheme~2 FEs.

\subsection{Interpretable score prediction: Attention-based MIL}
ABMIL~\cite{Ilse2018} has been adopted for several medical imaging tasks, such as screening of COVID-19 from chest computed tomography~\cite{Han2020}. MIL's capacity to account for within-image heterogeneity makes it well-suited for our task as RA-affected hands contain both healthy joints and joints with varying levels of damage, while the attention mechanism enables the model to selectively focus on disease-relevant regions. Our model uses the gated attention~\cite{Ilse2018}, which adopts hyperbolic tangent activation to avoid gradient explosion and sigmoid activation to introduce non-linearity. The patch features extracted by FEs are fed into the ABMIL model, scaled by attention weights, and summed to compute the image-level representation, which is subsequently passed through a regression FC layer for score prediction.

\section{Experiments and Results}

\subsection{Dataset, Implementation, and Evaluation}
\subsubsection{Dataset.}
This study used a public dataset of 3,818 hand radiographs collected from diagnosed and suspected RA patients (see an example in Fig.~\ref{fig_pipelines_overview}(A))~\cite{Wang2022}. The average SvdH scores, reported by two experienced radiologists, were provided with most samples representing early-stage RA (see Fig.~\ref{fig_pipelines_overview}(B)). The dataset was split into training, validation, and test sets, containing 2700, 760, and 358 images respectively. Of these, 143, 44, and 24 images were selected for nnU-Net foreground segmentation model development, and 1019, 334, and 153 images were selected for PC development. Using BoneFinder~\cite{Lindner2015}, we annotated 351 images for joint localisation model development, consisting of 245 training, 56 validation, and 50 test images.

\subsubsection{Implementation details.}
The SvdH scores were standardised to have a mean of 0 and a standard deviation (SD) of 1. For both schemes, random flip, intensity scaling (0.9 - 1.1), and rotation by 0\textdegree, 90\textdegree, 180\textdegree, or 270\textdegree\ were applied to the training images for augmentation. Random affine transformations (small-angle rotation, translation, and scaling only for joint localisation) were additionally applied when training Scheme 2 models, as the joint patching step could exclude the augmentation-induced artefacts. Random Gaussian noise, based on absolute radial distance, was added to the landmark predictions when cropping patches in the training set, using landmark-wise mean radial error (MRE) as SD. Whole images and patches were resized to $1024 \times 1024$ and $224 \times 224$ respectively, and normalised with the mean and SD of the training set. The PCs were initialised with ImageNet-pretrained model parameters. For Scheme 1 PCs, patches were cropped first and then augmented. For Scheme 2 PCs, images were augmented first, followed by joint patching. The dropout rate was set to 0.1 for ABMIL models. nnU-Net was trained with default configurations~\cite{Isensee2020}. Joint localisation models were trained using the setup in \cite{Viriyasaranon2023} for 300, 450, and 600 epochs with a batch size of 4 or 16. The pixel spacing in each image was estimated by assuming an average wrist width (distance between landmarks at the endpoints of a wrist) of 50 mm. Other models were trained using the stochastic gradient descent optimiser with a learning rate of 0.001, weight decay of 0.001, and momentum of 0.9. We chose mean squared error loss for score prediction and cross-entropy loss for patch classification. For Scheme 1 ABMIL models, we experimented with an input patch number $K$ of 30, 40, or 50. PCs and ABMIL models were initially trained for 100 epochs with a batch size of 4 or 16. An extra 50 or 100 epochs were added for models with slower convergence.
The code and newly created dual-hand landmark annotations will be available at: \url{https://github.com/ZhiyanBo/RA-AWMIL}.

\subsubsection{Evaluation metrics.}
Dice score was used to evaluate foreground segmentation models and classification accuracy to evaluate PCs. For joint localisation, MRE in mm and successful detection rate (SDR, \%) under 2, 3, 4, and 10 mm conditions were used. For score prediction, Pearson’s correlation coefficient (PCC), mean absolute error (MAE), and root mean squared error (RMSE) were used.

\subsection{Results and Discussion}

\subsubsection{Scheme 1: Patch classification.}
The nnU-Net achieved an average Dice score of 0.986 in 24 test images. Morphological operation-based post-processing further improved the quality of masks for noisy images, though in some cases fingertips were excluded. The MobileNetV2-based, ResNet-34-based, and ResNet-50-based PCs achieved 89.7\%, 89.0\%, and 90.0\% accuracy in the weakly supervised task. They successfully classified over 98\% of background patches while the misclassification rate was notably higher between normal and abnormal patches, as expected, since abnormal images may contain normal patches (even in severe RA, not all joints could be affected) and vice versa.

\begin{table}[!b]
\centering
\caption{Joint localisation performance of JLval and JLnoVal models in 1) all images involved in testing and 2) the test set after excluding images with poor-quality patches.}\label{tab_JL_models}
\resizebox{0.8\textwidth}{!}{%
\begin{tabular}{lllllcccc}
\toprule
\multirow{2.4}{*}{Dataset} & \multirow{2.4}{*}{\begin{tabular}[c]{@{}l@{}}Implementation\\ details\end{tabular}} & \multirow{2.4}{*}{\begin{tabular}[c]{@{}l@{}}Epoch \\ number\end{tabular}} & \multirow{2.4}{*}{\begin{tabular}[c]{@{}l@{}}MRE\\ (mm, SD)\end{tabular}} & \multicolumn{4}{c}{SDR (\%)} \\ \cmidrule{5-8}
 &  &  &  & \multicolumn{1}{c}{2 mm} & \multicolumn{1}{c}{3 mm} & \multicolumn{1}{c}{4 mm} & 10 mm \\ \midrule
\multirow{2}{*}{\begin{tabular}[c]{@{}l@{}}All images \\ (in test set/test \& validation sets)\end{tabular}} & JLval & 300 & 9.10 (40.14) & \multicolumn{1}{c}{88.89} & \multicolumn{1}{c}{92.89} & \multicolumn{1}{c}{94.16} & 95.27 \\ 
 & JLnoVal & 600 & 8.11 (38.04) & \multicolumn{1}{c}{88.37} & \multicolumn{1}{c}{92.96} & \multicolumn{1}{c}{94.72} & 95.93 \\ \midrule
\multirow{2}{*}{\begin{tabular}[c]{@{}l@{}}Images with high-quality patches \\ (in test set)\end{tabular}} & JLval & 300 & 0.88 (2.88) & \multicolumn{1}{c}{93.60} & \multicolumn{1}{c}{97.71} & \multicolumn{1}{c}{98.91} & 99.97 \\ 
 & JLnoVal & 600 & 0.85 (1.63) & \multicolumn{1}{c}{93.63} & \multicolumn{1}{c}{97.75} & \multicolumn{1}{c}{99.04} & 99.97 \\ \bottomrule
\end{tabular}%
}
\end{table}

\subsubsection{Scheme 2: Joint localisation and patch classification.}
We selected multiresolution HTC models with best testing performance under two cases: 1) JLval: among models with peak performance in the validation set during training, and 2) JLnoVal: among models extracted at the end of training.
Table \ref{tab_JL_models} shows the results in two cases: 1) all images used for testing, and 2) the test set after excluding images with poor-quality patches. Despite achieving over 94\% SDR within 4 mm, both models have high MREs with large SDs. We visually examined the images with high MRE and found that in images that are horizontally flipped (i.e., right hand on the left side), models were confused by the \textbf{R} label, which indicates the true \textbf{right} side. Although we labelled the landmarks based on their positions in the image, the models learned to differentiate between the true left and right hands. While this confusion caused severe MRE, these mistakes did not affect patch quality. However, because the model averages the predictions from two heatmaps of different resolutions, some predictions in flipped images were placed between the hands as one heatmap predicted based on the hands' relative position in the image, while one predicted based on the hands' true side. These errors led to wrong regions being cropped in 2 or 3 out of 50 images in the test set. Our results demonstrate the negative effect of anatomical symmetry on the landmark detection model's performance. To estimate the amount of error our joint patching scheme could tolerate, the landmark-wise MRE was calculated using the test set while excluding 4 or 5 images with symmetry-induced errors described above. With joint patches as inputs, the MobileNetV2-based, ResNet-34-based, and ResNet-50-based PCs achieved 92.3\%, 93.0\%, and 92.1\% accuracy with JLval and 91.8\%, 93.1\%, and 92.4\% accuracy with JLnoVal, suggesting they have successfully learned useful features of RA-related damage.

\subsubsection{SvdH score prediction.}
The results of SvdH score prediction are provided in Table \ref{tab_SvdH_pred}. With Scheme 1, the best performance was obtained when sampling 50 patches as inputs and fine-tuning the FE during training. The model with ResNet-34-based FE achieved a PCC of 0.932, MAE of 11.95, and RMSE of 17.16. Consistent performance improvement was observed when switching to Scheme 2 patches, potentially owing to further exclusion of anatomical structures and information irrelevant to SvdH scoring. With JLval joint localisation model, the best model JLval-ResNet50 used a ResNet-50-based FE, while with JLnoVal joint localisation model, the best model JLnoVal-MobileNetV2 used a MobileNetV2-based FE. Balancing all metrics, JLval-ResNet50 was selected as the best individual SvdH scoring model, achieving a PCC of 0.943, MAE of 11.33, and RMSE of 15.73. Ensembling JLval-ResNet50 and JLnoVal-MobileNetV2 by averaging their predictions yielded further performance improvement, reaching a PCC of 0.945, MAE of 11.22, and RMSE of 15.57. Both Scheme 1 and Scheme 2 models outperformed other published methods on dual-hand image-level SvdH score prediction in terms of MAE and RMSE~\cite{Bo2024,Moradmand2025,Wang2022}. Scheme 2 models performed better than Scheme 1 models in scoring moderate and late-stage cases, though their accuracy in early and moderate-stage cases could be further improved, as displayed in Fig.~\ref{fig_pipelines_overview}(D). Compared with experienced radiologists, JLval-ResNet50 demonstrated as good or even better average prediction accuracy, as shown by its lower MAE and RMSE. However, its PCC was 2.7\% lower, suggesting scope for pipeline optimisation.

\begin{table}[!t]
\centering
\caption{Pipeline performance in SvdH score prediction and the published inter-rater differences~\cite{Wang2022}. The top two values of a metric are made \textbf{bold} and \underline{underlined}.}\label{tab_SvdH_pred}
\resizebox{0.75\textwidth}{!}{
\begin{tabular}{lllccc}
\toprule
\multicolumn{1}{l}{\textbf{Method}} & \multicolumn{1}{l}{\textbf{Implementation details}} & \textbf{Feature extractor} & \textbf{PCC} & \textbf{MAE} & \textbf{RMSE} \\ \midrule
\multicolumn{1}{l}{\multirow{3}{*}{\textbf{\begin{tabular}[c]{@{}l@{}}Scheme 1:\\ RA-severity \\ based patches\end{tabular}}}} & \multicolumn{1}{l}{\multirow{3}{*}{\begin{tabular}[l]{@{}l@{}}50 patches as inputs\\ \&\\ Fine-tuned FE\end{tabular}}} & MobileNetV2 & 0.924 & {\ul 12.31} & 18.04 \\ 
\multicolumn{1}{l}{} & \multicolumn{1}{l}{} & ResNet-34 & \textbf{0.932} & \textbf{11.95} & \textbf{17.16} \\ 
\multicolumn{1}{l}{} & \multicolumn{1}{l}{} & ResNet-50 & {\ul 0.925} & 12.68 & {\ul 17.88} \\ \midrule
\multicolumn{1}{c}{\multirow{7}{*}{\textbf{\begin{tabular}[c]{@{}l@{}}Scheme 2: \\ Joint patches\end{tabular}}}} & \multicolumn{1}{l}{\multirow{3}{*}{JLval}} & MobileNetV2 & 0.938 & 11.55 & 16.50 \\ 
\multicolumn{1}{l}{} & \multicolumn{1}{l}{} & ResNet-34 & 0.938 & 11.46 & 16.37 \\ 
\multicolumn{1}{l}{} & \multicolumn{1}{l}{} & ResNet-50 & \textbf{0.943} & {\ul 11.33} & \textbf{15.73} \\ \noalign{\vskip 0.5ex}
\cdashline{2-6}[2pt/2pt] \noalign{\vskip 0.9ex}
\multicolumn{1}{l}{} & \multicolumn{1}{l}{\multirow{3}{*}{JLnoVal}} & MobileNetV2 & {\ul 0.941} & 11.61 & {\ul 16.23} \\ 
\multicolumn{1}{l}{} & \multicolumn{1}{l}{} & ResNet-34 & 0.938 & \textbf{11.29} & 16.53 \\ 
\multicolumn{1}{l}{} & \multicolumn{1}{l}{} & ResNet-50 & 0.938 & 11.88 & 16.37 \\ \noalign{\vskip 0.5ex} \cdashline{2-6}[2pt/2pt] \noalign{\vskip 0.9ex}
\multicolumn{1}{c}{} & \multicolumn{2}{l}{Ensemble of the best models} & \textbf{0.945} & \textbf{11.22} & \textbf{15.57} \\ \midrule
\multicolumn{1}{l}{\multirow{3}{*}{\textbf{\begin{tabular}[c]{@{}l@{}}Published \\ methodologies\end{tabular}}}} & \multicolumn{2}{l}{ResNet-Dwise50~\cite{Wang2022}} & 0.97 & 14.90 & 22.01 \\ 
\multicolumn{1}{l}{} & \multicolumn{2}{l}{Ensemble of CNNs~\cite{Bo2024}} & 0.925 & 12.57 & 18.02 \\ 
\multicolumn{1}{l}{} & \multicolumn{2}{l}{Custom ViT~\cite{Moradmand2025}} & \xmark & 21.14 & 44.28 \\ \midrule
\multicolumn{3}{l}{\textbf{Published inter-rater differences~\cite{Wang2022}}} & 0.97 & 12.24 & 18.75 \\ \bottomrule
\end{tabular}%
}
\end{table}

\subsubsection{Attention maps for interpretable RA scoring.}
\begin{figure}[!t]
\centering
\includegraphics[width=1\textwidth]{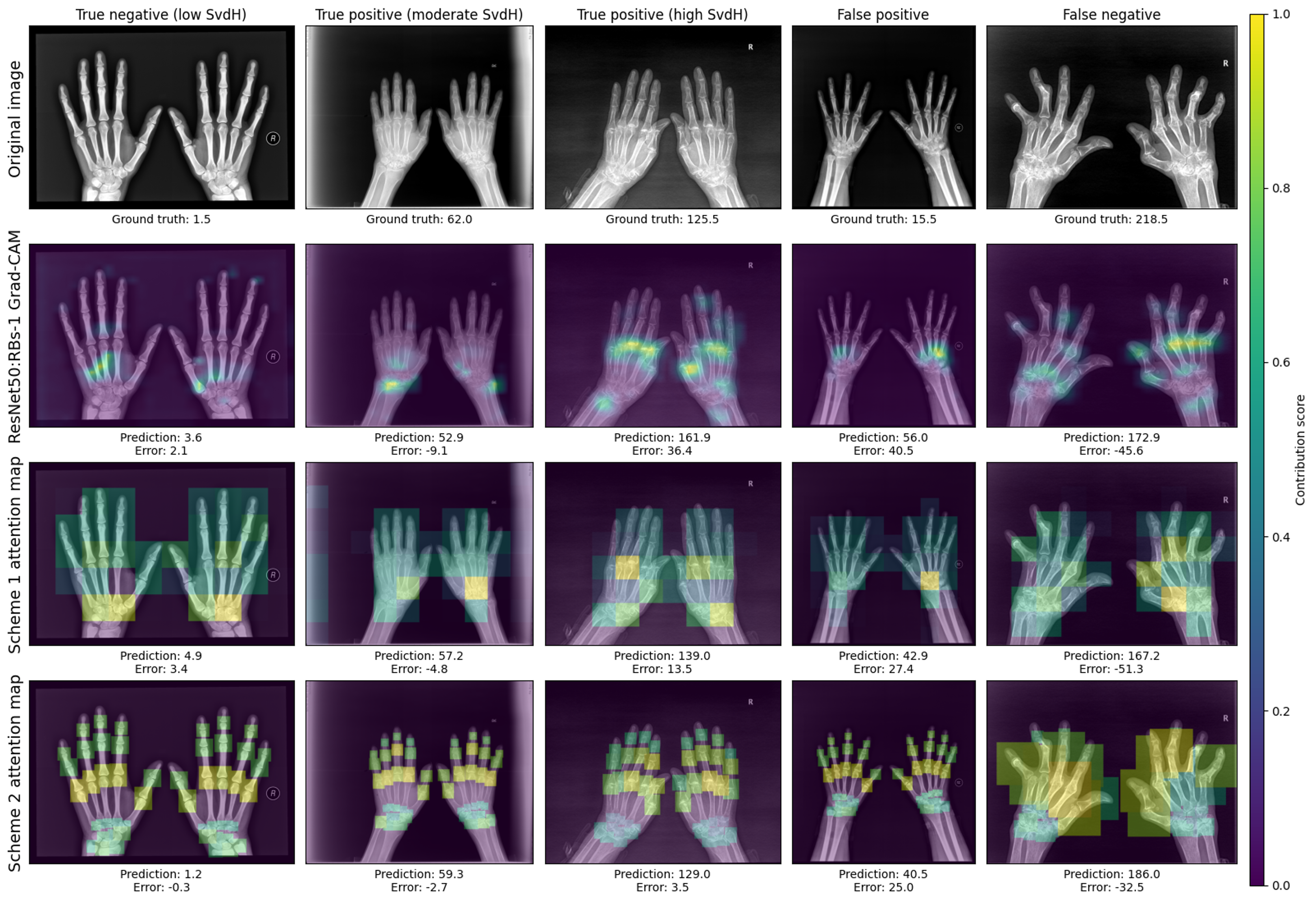}
\caption{Grad-CAM and attention maps of example images from the best individual whole-image model (ResNet-50:RBs-1) \cite{Bo2024}, and our Scheme 1 and Scheme 2 models.} \label{fig_attention_maps}
\end{figure}

Fig.~\ref{fig_attention_maps} displays the Grad-CAM heatmaps of ResNet-50:RBs-1, the SOTA individual single-stage model from our previous work~\cite{Bo2024} that takes a whole image as input, and the attention maps of our models for images with varying RA severity and prediction accuracy. An experienced rheumatologist was invited to review the visual explanations, who noted that ResNet-50:RBs-1 could incorrectly focus on RA-irrelevant structures. Capable of identifying some but not all damage in each example, it wrongly highlighted diaphyses and arm bones in true negative and true positives (TPs). Scheme~1 performed better in detecting wrist damage, but failed to highlight finger damage in TPs. Also, some joints were split into multiple patches, causing information loss. On the other hand, Scheme~2 offered better interpretability than Scheme~1 as its patches were more specific and included larger proportions of RA-relevant structures. However, joint localisation accuracy may deteriorate in images with curved fingers, leading to suboptimal patches as in the false negative example and subsequently large prediction errors. Best at detecting finger damage, Scheme~2 could over-focus on finger joints, especially the metacarpophalangeal joints, while not highlighting wrist damage enough, as in false positive and TPs. For clarification, this clinician evaluation is based on a small number of images, which may not cover the full spectrum of model behaviours.

\section{Conclusion}
This work presents an interpretable anatomy-aware ABMIL framework for image-level RA quantification by SvdH scoring, achieving SOTA performance and accuracy comparable to experienced radiologists. Our two RA-relevant region sampling strategies effectively capture damaged features, with joint patches acquired using a multiresolution HTC further enhancing model performance and interpretability. Our work shows that by incorporating prior knowledge of the disease, ABMIL could generate better image-level representations, further boosting prediction accuracy. Future work will focus on improving joint localisation robustness through noise removal and enhancing prediction performance, particularly in early-stage cases. A structured clinician evaluation study would also provide insights into the perceived accuracy and usefulness of the pipeline.

\noindent\textbf{Prospect of Application}: Our SvdH scoring model has the potential to serve as an additional rater in clinical trials, offering more consistent trial endpoints by reducing inter-rater variability. Furthermore, by providing anatomy-aware visual explanations, it may enhance clinician trust and enable objective quantification of joint damage for the diagnosis and monitoring of RA.

\begin{credits}
\subsubsection{\ackname}
This work was supported by the EPSRC Centre for Doctoral Training in Health Data Science (EP/S02428X/1). We would like to thank the Oxford Biomedical Research Computing Facility for providing the computational resources.

\subsubsection{\discintname} 
The authors have no competing interests to declare that are relevant to the content of this article.

\end{credits}

\bibliographystyle{splncs04}
\bibliography{AMAI2025_15}

\end{document}